\documentclass[aps,prl,twocolumn,amsmath,amssymb,nofootinbib,superscriptaddress,floatfix]{revtex4-1}

\usepackage{times}
\usepackage[pdftex]{graphicx}
\usepackage{dcolumn}
\usepackage{bm}
\usepackage{amsmath}
\usepackage{indentfirst}
\usepackage{float}
\usepackage[colorlinks]{hyperref}
\usepackage[dvipsnames]{xcolor}
\usepackage{multirow}
\usepackage{verbatim}

\begin{document}

\title{Density peak clustering using tensor network}

\author{Xiao Shi}
\affiliation{Institute of Mathematics, Academy of Mathematics and Systems Science, Chinese Academy of Sciences, Beijing
100190, China}
\affiliation{School of Mathematical Sciences, University of Chinese Academy of Sciences, Beijing 100049, China}

\author{Yun Shang}
\email{shangyun@amss.ac.cn}
\affiliation{Institute of Mathematics, Academy of Mathematics and Systems Science, Chinese Academy of Sciences, Beijing
100190, China}
\affiliation{NCMIS, MDIS, Academy of Mathematics and Systems Science, Chinese Academy of Sciences, Beijing, 100190,China}

\begin{abstract}
Tensor networks, which have been traditionally used to simulate many-body physics, have recently gained significant attention in the field of machine learning due to their powerful representation capabilities. In this work, we propose a density-based clustering algorithm inspired by tensor networks. We encode classical data into tensor network states on an extended Hilbert space and train the tensor network states to capture the features of the clusters. Here, we define density and related concepts in terms of fidelity, rather than using a classical distance measure. We evaluate the performance of our algorithm on six synthetic data sets, four real world data sets, and three commonly used computer vision data sets. The results demonstrate that our method provides state-of-the-art performance on several synthetic data sets and real world data sets, even when the number of clusters is unknown. Additionally, our algorithm performs competitively with state-of-the-art algorithms on the MNIST, USPS, and Fashion-MNIST image data sets. These findings reveal the potential of tensor networks for machine learning applications.
\end{abstract}

\date{\today}
\maketitle

\vspace{8mm}

\section{1.Introduction}
Clustering is a fundamental problem in unsupervised learning, with numerous applications in various fields such as natural~\cite{zhu2022flexible, PhysRevB.71.201309,murtagh2012algorithms, min2018survey, madhulatha2012overview} and social sciences~\cite{cook2005emergent,klerkx2019review}. The goal of clustering algorithms is to divide a data set into different clusters according to a certain standard, such that the similarity within each cluster is maximized while the difference between clusters is minimized. Existing clustering algorithms can be mainly divided into partition-based methods~\cite{10.14778/2078331.2078340, mazzeo2017fast}, hierarchical methods~\cite{johnson1967hierarchical,chen2007efficient} and density-based clustering~\cite{campello2020density,lv2016efficient}. The partition-based clustering methods, such as K-means~\cite{MacQueen1967}, CLARA~\cite{KAUFMAN1986425}, and PCM~\cite{rdusseeun1987clustering}, divide data points into different categories based on certain algorithm rules. However, these methods require prior knowledge of the number of clusters or cluster centers. In contrast, hierarchical methods, such as AGNES~\cite{barbakh2009review} and BIRCH~\cite{zhang1996birch}, decompose a given data set hierarchically until a certain condition is met. They can be divided into two schemes: "bottom-up" and "top-down." On the other hand, density-based clustering algorithms, such as DBSCAN~\cite{10.5555/3001460.3001507}, OPTICS~\cite{ankerst1999optics}, and DPC~\cite{rodriguez2014clustering}, examine the connectivity between samples from the perspective of sample density and continuously expand clusters based on the connectable samples to obtain the final clustering results. These algorithms are capable of finding clusters of all shapes and sizes in noisy data, without the need for prior knowledge of the number of clusters. However, in the era of information explosion, how to manage massive data is one of the challenges people often confront~\cite{bai2017fast,LV20169}, which means that machine learning requires novel techniques to efficiently manage broad and diverse sets of data.

 Quantum Machine Learning (QML) is an emerging field of research that combines the principles of quantum computing and classical machine learning. By leveraging the unique properties of quantum mechanics, such as quantum coherence and quantum entanglement~\cite{RevModPhys.89.041003,RevModPhys.91.025001,PhysRevLett.78.2275}, QML aims to develop new algorithms that can speed up or optimize classical machine learning tasks. Several quantum machine learning algorithms have been proposed, which achieves an exponential speedup compared to the classical algorithm~\cite{lloyd2014quantum,PhysRevLett.113.130503,lloyd2013quantum,PhysRevLett.114.110504,liu2021rigorous}. Lloyd~\cite{lloyd2013quantum} proposed the quantum K-means algorithm, which achieves an exponential speedup compared to the classical algorithm. Cai~\cite{PhysRevLett.114.110504} verified the algorithm with 4, 6, and 8 qubit scales on a small optical quantum computer. In \cite{liu2021rigorous}, the authors investigated the potential of quantum kernel methods for machine learning tasks. They demonstrated that quantum kernel methods can provide significant quantum speedups while maintaining high accuracy and robustness compared to conventional kernel methods in supervised classification tasks. 

However, despite its potential, QML faces several challenges such as the limited number of physical qubits available and high error rates in near-term devices. Tensor network states, which are powerful numerical tools for studying quantum many-body systems, have been used to solve classical machine learning problems and achieve outstanding results~\cite{PhysRevLett.88.018702,PhysRevA.105.052424,PhysRevLett.103.160601,PhysRevLett.115.180405,PhysRevLett.126.170603}. Tensor networks are a powerful mathematical framework for representing and manipulating high-dimensional data. They have been extensively studied in the fields of quantum physics and quantum information science, and include representations such as Matrix Product States (MPS), Projected Entangled Pair States (PEPS), and Multi-scale Entanglement Renormalization Ansatz (MERA)\cite{verstraete2004renormalization, PhysRevLett.69.2863, PhysRevLett.101.110501,PhysRevA.75.033605}. Among these, the MPS is one of the most widely used representations for quantum states. Originally proposed in the context of the density matrix renormalization group, MPS decompose a rank-N tensor into a chain of rank-3 tensors of length N, providing an efficient, natural, and compact representation for high-dimensional quantum states via low-rank approximation. In particular, MPS have been utilized to simulate the ground and excited states of infinite one-dimensional quantum systems with great success\cite{PhysRevB.94.165116,SCHOLLWOCK201196}. Also, in recent years, MPS have demonstrated good performance on machine learning tasks~\cite{stoudenmire2016supervised,PhysRevX.8.031012,PhysRevB.101.075135}.

Now, most of the existing clustering algorithms assume prior knowledge of the number of clusters. However, it is often unknown in the real world. In this paper, we present a novel density-based clustering algorithm that utilizes MPS to automatically identify the number of clusters. Our algorithm encodes the data into MPS, leveraging quantum entanglement to create a higher-dimensional representation of the data, which allows for more expressive modeling. Our proposed algorithm is based on the principles of Density-Peak Clustering (DPC)~\cite{rodriguez2014clustering} but utilizes MPS to capture data in a higher-dimensional Hilbert space. This allows for a more expressive representation of the data and facilitates the identification of a hyperplane that can linearly separate the data, even if it is not linearly separable in the original space. Additionally, key properties such as density, core point, and border point are redefined using fidelity measures to achieve improved clustering results. Empirical evaluations on commonly used data sets demonstrate the algorithm's excellent efficiency and accuracy, even when the number of clusters is unknown.

The rest of this paper is organized as follows. Section 2 provides an overview of common density-based clustering algorithms and tensor network machine learning algorithms. In Section 3, we introduce our proposed density-based tensor network algorithm. Numerical simulation experiments and results analysis are presented in Section 4. Finally, a summary and outlook for future work are stated in Section 5.

\section{2.Related work}
\subsection{2.1. Density-based clustering algorithm}
In contrast to K-means algorithms that cannot cluster non-convex data sets, density-based clustering algorithms can find clusters of arbitrary shape and size in the data. Usually, density clustering starts from the perspective of data point density, considers the connectivity between different samples, and continuously expands the cluster based on the connectable samples to obtain the final clustering results. One of the most 
well-known algorithms is the DBSCAN algorithm. Its main idea is to first find data points with higher density, and then gradually connect adjacent high-density points together to generate different clusters. As one of the earliest proposed density-based clustering algorithms, inspired by this, many improved versions have been developed, such as GDBSCAN~\cite{sander1998density}, Recon-DBSCAN\cite{ZHU2016983}, RNN-DBSCAN~\cite{8240674}, OPTICS~\cite{ankerst1999optics}.

DPC algorithm, as a new type of density clustering algorithm proposed in 2014, is based on two intuitive assumptions: 1) the local density of data points close to the cluster center is relatively low; 2) the distance between any cluster center and the data points with higher density is far. In order to complete the clustering task, for any data point $x_i$, DPC needs to calculate 2 quantities: the local density $\rho_i$ of $x_i$ and its distance to the closest point with higher density $\delta_i$. Let $ d_{ij}$ represents the distance between points $x_i$ and $x_j$, $d_c$ is a predefined cut-off distance. The definition of $\rho_i$ is given as:
\begin{align}
\rho_i=\mathop{\sum} \limits_{j}\chi(d_{ij}-d_c)
\end{align}

where
\begin{align}
\chi(x)=\left\{
\begin{aligned}
&1&x<0\\
&0&otherwise \\
\end{aligned}
\right.
\end{align}

In addition to the density method of formula 1, when the data set is small, the Gaussian kernel is usually used to calculate the density of continuous points, which is defined as follows:
\begin{align}
\rho_i=\mathop{\sum}\limits_j exp(-\frac{d_{ij}^2}{d_c^2})
\end{align}

$\delta_i$ is then defined by
\begin{align}
\delta_i = \left\{\begin{aligned}
&\mathop{min} \limits_{j:\rho_i<\rho_j}(d_{ij})&\rho_i\neq max(\rho)\\
&\mathop{max} \limits_{j}(d_{ij})&otherwise \\
\end{aligned}\right.
\end{align}

After calculating these two values, the point with the larger $\rho_i$ and $\delta_i$ is selected as the cluster center by drawing decision graph. Then assign the remaining points to the nearest neighbor cluster centers. Although the DPC algorithm is a simple and effective algorithm, but it also has some drawbacks. For example, there are two ways to define density for large data sets and small data sets, but there is no objective indicator to define whether the data set is large or small. And it only classifies data according to Euclidean distance, which will results in "chain reaction". In recent years, a series of new algorithms have been proposed to solve these problems. For example, KNN-DPC~\cite{DU2016135} optimizes the density function, and DPC-CE~\cite{guo2022density} optimizes DPC by introducing a graph-based connectivity estimation strategy, etc. 

\subsection{2.2. Tensor Network Machine Learning Algorithms}

Recently, the tensor network method has become an important theoretical and computational tool in the fields of classical statistics and quantum many-body physics~\cite{PhysRevLett.69.2863, SCHOLLWOCK201196,RevModPhys.93.045003}. At the same time, machine learning inspired by tensor networks becomes an emerging topic. Tensor networks are able to capture the underlying structures and patterns in complex data by efficiently representing high-dimensional tensors using a much lower number of parameters. This makes them well-suited for tasks such as data compression, feature extraction, and dimensionality reduction, which are all important tasks in machine learning. Since tensor networks can map precisely to quantum circuits, one exciting line of research in tensor network machine learning is to deploy and even train such models on quantum hardware~\cite{Huggins_2019, cong2019quantum}. At the same time, since tensor networks can learn probability distributions for given data, this also naturally bridges quantum many-body physics and machine learning.

Tensor network can be combined with machine learning in two ways. One is to directly use tensor network as machine learning model architecture. The other is to use tensor networks to compress layers or other auxiliary tasks in the neural network architecture. In ~\cite{stoudenmire2016supervised}, MPS have been used as proof-of-principle for tensor networks in supervised learning. In ~\cite{PhysRevX.8.031012}, MPS have been used to parameterize generative models, and update the tensors in MPS by gradient descent. Sun et al.~\cite{PhysRevB.101.075135} implemented the classification task with MPS, and its results exceeded a series of baseline models such as naive Bayes classifiers, SVMs, etc. Gao et al.~\cite{PhysRevResearch.2.023300} replaced the fully connected layers in the neural network with MPO, which greatly compresses the parameters without affecting the accuracy. By combining with some good initialization methods, Shi et al. ~\cite{PhysRevA.105.052424} proposed to use tensor network to solve the clustering problem, and they achieved the best clustering results on some data sets.

\section{3.Tensor Network Clustering Algorithms}

\subsection{3.1 Training MPS}

In order to train an MPS with all the data, the first thing is to map all the data into quantum states. The $i$-th element $x_i^n$ of the input vector $\vec{x^n}=(x_1^n, x_2^n,...,x_m^n)$ of length m is mapped to a superposition of quantum states $|0\rangle$ and $|1\rangle$,which can be described as

\begin{align}
    |\psi_i^n\rangle = cos(\frac{\pi}{2}x_i^n)|0\rangle+ sin(\frac{\pi}{2}x_i^n)|1\rangle
\end{align}

Therefore, the input vector $\vec{x^n}$ can be written as the tensor product of $|\psi_i^n\rangle$

\begin{align}
\Psi(\vec{x^n})= |\psi_1^n\rangle\otimes|\psi_2^n\rangle\otimes...\otimes|\psi_m^n\rangle
\end{align}

where it can be expressed in the form of a tensor network as

\begin{align}
\Psi_{i}^{\vec\sigma} = \mathop{\sum} \limits_{\alpha_0,\alpha_1,...,\alpha_m}X^{\sigma_0}_{i,\alpha_0,\alpha_1}X^{\sigma_1}_{i,\alpha_1,\alpha_2...}X^{\sigma_{m-1}}_{i,\alpha_{m-1},\alpha_m}
\end{align}

Where $\sigma_i$ is its physical indices, $\alpha_i$ is its auxiliary indices with $\alpha_0=\alpha_1=...=\alpha_n=1$. Each $X_{i}$ represents a $1\times 2 \times1$ third-order tensor whose elements are

\begin{align}
X_{i,1,1}^0=cos(\frac{\pi}{2}x_{i,l}), X_{i,1,1}^1=sin(\frac{\pi} {2}x_{i,l})
\end{align}

Analogously, we randomly generate a quantum state $\Phi^{\vec\tau}$ of length $m$ and bond dimension equals to D in the form of MPS:

\begin{align}
\Phi^{\vec\tau} = \mathop{\sum} \limits_{\beta_0,\beta_1,...,\beta_m}Y^{\tau_1}_{\beta_0,\beta_1}Y^{ \tau_1}_{\beta_1,\beta_2...}Y^{\tau_n}_{\beta_{m-1},\beta_m}
\end{align}

Where $\beta$ is an auxiliary indices, which determines the upper limit of the entanglement entropy that this MPS state can accommodate, and $1\leq\beta_i\leq D$. After completing the above steps, a variational matrix product states algorithm will be used to update the parameters in the MPS. Specifically, the randomly initialized quantum state $\Phi^{\vec\tau}$ is converted into canonical forms, and the NLL function is adopted as the loss function

\begin{align}
f(\Phi^{\vec\tau})=ln|\langle\Phi^{\vec\tau\dagger}|\Phi^{\vec\tau}\rangle|-\frac{1}{| \Gamma|}\mathop{\sum}\limits_{i}\ln|\Phi^{\vec\tau\dagger}\Psi_{i}^{\vec\sigma}|^2
\end{align}

Where $|\Gamma|$ is the number of data points. Gradient descent is used to update each tensor when the quantum state $\Phi^{\vec\tau}$  satisfies the normalization condition:

\begin{align}
Y^{\tau_i}-\eta\frac{\partial f}{\partial Y^{\tau_i}}\rightarrow Y^{\tau_i}
\end{align}

where $\eta$ is the learning rate. Tensors will be updated from the first to the last, and then back, this process is called a sweep. The iteration process will stop when the sweep reaches the maximum number of iterations or the loss function converges. After this, the quantum state $\Phi^{\vec\tau}$ will give the joint distribution probability of pixels.

\subsection{3.2 Generation of clusters}

After training through the algorithm in the last section, a quantum state $\Phi^{\vec\tau}$ with bond dimension equal to D is obtained. Similar to the DPC algorithm, we define $\rho$ and $\delta$ by fidelity measure instead of distance measure, and replace the Gaussian kernel function with the Sigmoid kernel function:

\begin{align}
\rho_i=tanh(f_{i}/(10*f_c))
\end{align}

where $f_i=|\langle\Psi_i^{\vec\sigma} | \Phi^{\vec\tau}\rangle |$ represents the fidelity between quantum state $\Psi_i^{\vec\sigma}$ and $\Phi^{\vec{\tau}}$, $f_c$ is a predefined cutoff distance. Similarly, computing the fidelity $f_{ij}=|\langle\Psi_i^{\vec\sigma}|\Psi_j^{\vec\sigma}\rangle|$ of the quantum states of the data points $x_i$ and $x_j$, we get the definition of $\delta_i$ as follows

\begin{align}
\delta_i=\mathop{min} \limits_{j:\rho_j>\rho_i}(f_{ij})
\end{align}

Different from the DPC algorithm, we adopt a method similar to Ref \cite{ren2020deep} here, and select the point with relatively large rho and delta as the local cluster center. The definition of local clusters is given in Def 1.\\

\textbf{Def 1: (Local Cluster Center)}

\emph{A data point $x_i$ is defined as a local cluster center if it satisfies the conditions $\delta_i>f_c$ and $\rho_i>\overline\rho$, where $\overline\rho$ is the average of all point density.}\\

After the local cluster centers have been identified, the remaining points can be assigned to their nearest higher-density neighbor to generate a set of local clusters. Therefore, L local clusters $(C^{(1)}, C^{(2)},...,C^{(L)})$ are obtained.

Similar to the DBSCAN algorithm, for each local cluster $C^{(k)}$, core point and border point need to be defined. Therefore, for the local cluster $C^{(k)}$, an MPS representation  $\Phi^{\vec\tau}_{k}$ is trained using the method in Section 3.1. The definitions of core point and border point are given in Def 2.\\

\textbf{Def 2: (Definition of core point and border point)}

\emph{Assuming that the point $c_i$ in the local cluster $C^k$ satisfies $f_i'>\overline f_k'$, where $f_i'= |\langle \Psi_{i}^{\vec\sigma}|\Phi^{\vec\tau}_{k}\rangle|, \overline f_k' = \frac{1}{n_k}\sum\limits_{c_j\in C^{(k)}} |\langle \Psi_{j}^{\vec\sigma}|\Phi^{\vec\tau}_{k}\rangle|$, and $n_k$ is the number of points in the local cluster $C^k$. Then the point $c_i$ is called the core point of the local cluster $C^k$, otherwise $c_i$ is the border point.}\\


In the following, in order to determine the connectivity between local clusters, we will give the definitions of Density Directly-connectable and Density Connectable respectively.\\

\textbf{Def 3: (Density Directly-connectable of Clusters)}

\emph{If there are core points $c^i\in C^i$, $c^j\in C^j$ in local clusters $C^{i}$ and $C^j$, and $f_{ij}<f_d $, where $f_d$ is a predefined parameter, then the local clusters $C^i$ and $C^j$ is directly-connectable.}\\

\textbf{Def 4: ((Density Connectable of Clusters)}

\emph{If there are paths $C^i,C^1,C^2,...,C^n,C^j$, where $C^i$ and $C^1$, $C^k$ and $C^{(k+1)}$, $C^n$ and $C^j$ are all Density Directly-connectable, then $C^i$ and $C^j$ are called Density Connectable.}\\

Finally, all local clusters with density connectable are merged to get the final clustering result. It is easy to verify that our algorithm does not require an input number of classes and can find clusters of arbitrary shapes.

\section{4.Experimental results and analysis}
\subsection{4.1 Experimental results on synthetic data sets}
Our first experiment is to apply the algorithm to six commonly used synthetic data sets: Twomoons, Jain, Threecircles, Smile, Fourlines, and Unbalance. Both Twomoons and Jain data sets consist of two moon-shaped clusters, but the size of the two data sets is different, and the density of the clusters is not the same. Some manifold data sets, such as Threecircles and Smile, can be used to evaluate the performance of the algorithm on non-spherical clusters. Fourline is represented as a linearly separable data set consisting of 4 linearly non-uniform density clusters. Unbalance is a large-scale synthetic data set consisting of multiple spherical clusters. Their main characteristics are summarized in Table ~\ref{tab:tab1}. Note that in the process of mapping data into quantum states, since the mapping function is a trigonometric function, we first scale the data between 0 and 1 using max-min normalization to avoid problems with periodicity. In comparative experiments, the proposed algorithm is compared with other methods such as K-means~\cite{MacQueen1967}, DPC~\cite{rodriguez2014clustering}, DBSCAN~\cite{10.5555/3001460.3001507}, SNN-DPC~\cite{LIU2018200}, DGDPC~\cite{zhang2021density}, DPC-CE~\cite{guo2022density}. Among them, K-means and DBSCAN are commonly used as classical clustering algorithms. The DPC algorithm is used as the benchmark algorithm. SNN-DPC, DGDPC and DPC-CE are three better revisions of DPC algorithm. The parameters that need to be pre-specified in all these algorithms are listed in Table ~\ref{tab:tab2}.

\begin{figure}
\includegraphics[width=\columnwidth]{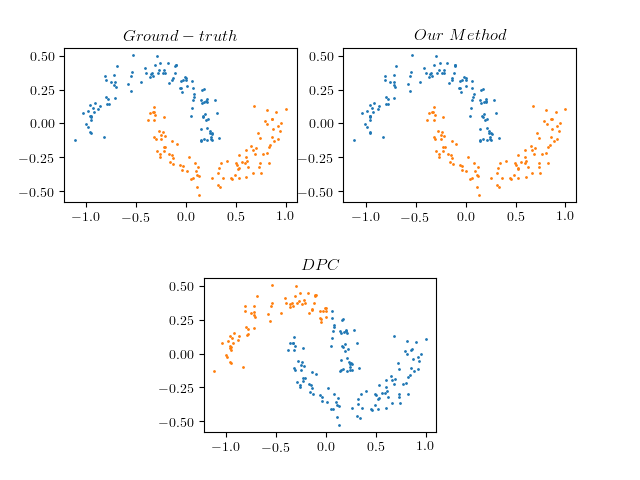} 
\caption{Compared results of DPC and our method on Twomoons data set.}
\label{fig:fig1} 
\end{figure}

\begin{figure}
\includegraphics[width=\columnwidth]{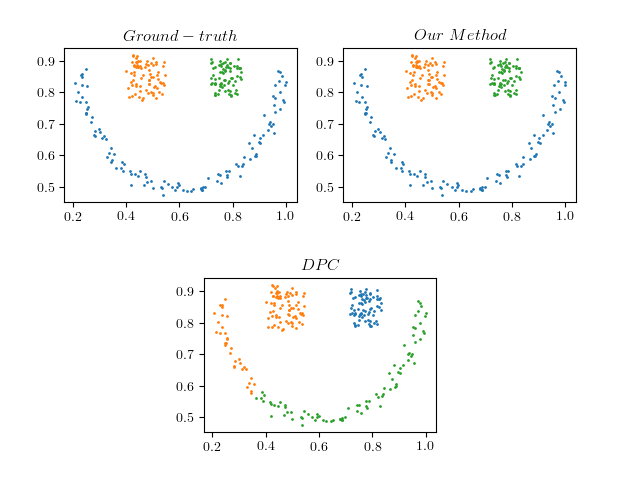} 
\caption{Compared results of DPC and our method on Smile data set.}
\label{fig:fig2} 
\end{figure}

\begin{figure}
\includegraphics[width=\columnwidth]{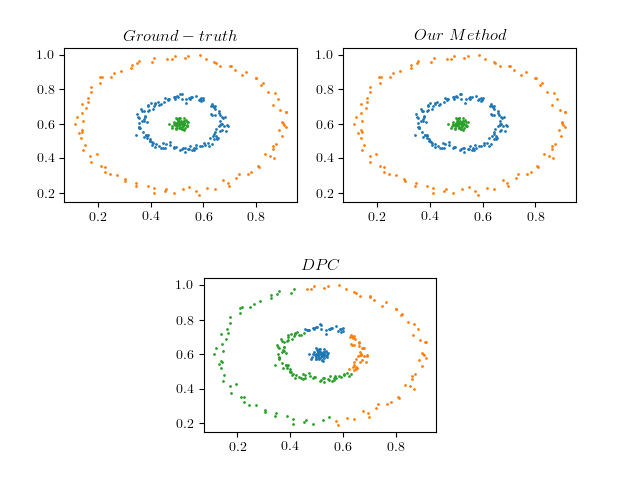} 
\caption{Compared results of DPC and our method on Threecircles data set.}
\label{fig:fig3} 
\end{figure}

\begin{figure}
\includegraphics[width=\columnwidth]{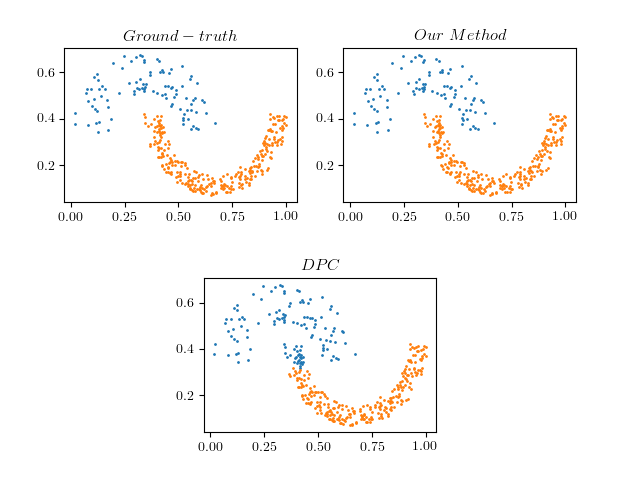} 
\caption{Compared results of DPC and our method on Jain data set.}
\label{fig:fig4} 
\end{figure}

\begin{figure}
\includegraphics[width=\columnwidth]{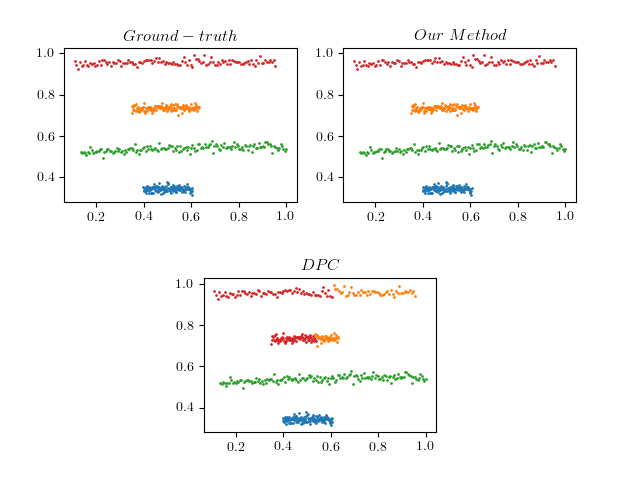} 
\caption{Compared results of DPC and our method on Fourlines data set.}
\label{fig:fig5} 
\end{figure}

\begin{figure}
\includegraphics[width=\columnwidth]{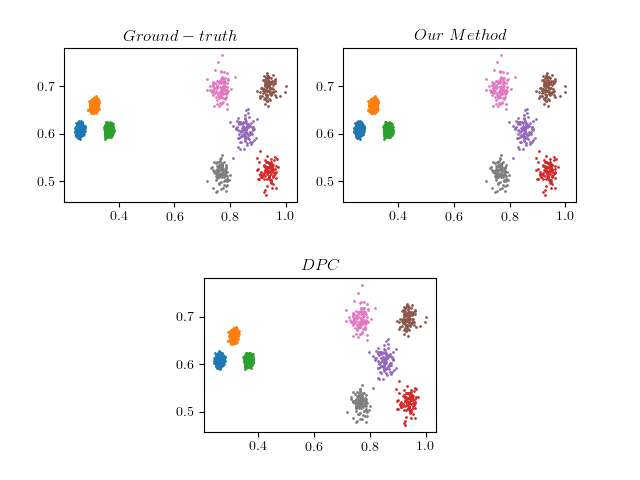} 
\caption{Compared results of DPC and our method on Unbalance data set.}
\label{fig:fig6} 
\end{figure}

Figure ~\ref{fig:fig1}-~\ref{fig:fig6} visualizes the differences between the DPC algorithm, our algorithm, and its true label. We use three kinds of popular external evaluation index of clustering algorithms called FMI, ARI and NMI to evaluate all the clustering results. Table ~\ref{tab:tab3} compares the effectiveness of these methods numerically. And in the iterative process of MPS, the upper limit of its sweeps is set to 30. It can be seen from the results that our method can achieve 100\% accuracy on 5 of the data sets. Significantly better than DBSCAN and DPC algorithms on Threecircle and Jain data sets. Our algorithm also outperforms DPC on the Twomoons, Smile and Fourlines data sets. 
Only on the Unbalance dataset, our algorithm is slightly lower than DPC and DBSCAN.

\begin{table}
\caption{Description of the synthetic data sets.} 
\label{tab:tab1}
\begin{tabular}{| c | c | c | c | }
\hline                              
Data set & Point & Attribute & Clusters \\
\hline
Twomoons & 200 & 2 & 2 \\
Smile & 266 & 2 & 3 \\
Threecircles & 299 & 2 & 3\\
Jain & 373 & 2 & 2 \\
Fourlines & 512 & 2 & 4 \\
Unbalance & 6500 & 2 & 8 \\
\hline
\end{tabular}
\end{table}

\begin{table}
\caption{Parameters configurations of compared algorithms and our method} 
\label{tab:tab2}
\begin{tabular}{| c | c | }
\hline                              
Algorithms &  Parameters setting \\
\hline
K-means & $cluster\ number\ k$\\
DPC & $d_c=1\% \sim 5\%$ \\
DBSCAN & $Eps=0.5 \sim 3,MinPts=4$ \\
SNN-DPC & $d_c=2\%\ \sim 3\%,3\leq K\leq 50$ \\
DGDPC & $d_c=1\% \sim 5\%, m=0.1\sim 1$\\
DPC-CE & $d_c=2\%,T_r=0.25,P_r=0.3$\\
\hline
Our method & $d_c=0.05\%  \sim 0.6\%,f_d = 0.93 \sim 0.999,D=8$\\
\hline
\end{tabular}
\end{table}

\begin{table}[!htbp]
\caption{Clustering results of all methods in synthetic data set, where in our method, MPS bond is set to 8.}
\label{tab:tab3}
\centering
\begin{tabular}{ | c | c | c | c | c | c | c |  }
\hline

\multicolumn{1}{|c|}{\multirow{2}*{Method}}&\multicolumn{3}{c|}{Twomoons}&\multicolumn{3}{c|}{Smile}\\
\cline{2-7}
~ & FMI & ARI & NMI & FMI & ARI & NMI
\\
\hline
K-means    &    0.5683 & 0.1401    &    0.1077 & 0.6155  & 0.4022 & 0.5318\\
\hline
DPC    &    0.7175 & 0.4068    &   0.4584 & 0.7627  & 0.5264 & 0.7080\\
\hline
DBSCAN~    &    \textbf{1.0}    &   \textbf{1.0}   &   \textbf{1.0}    &   \textbf{1.0}    &  \textbf{1.0}    &  \textbf{1.0} \\
\hline
SNN-DPC    &    0.7175    &   0.4068   &   0.4587    &   \textbf{1.0}    &  \textbf{1.0}    &  \textbf{1.0} \\
\hline
DGDPC    &    \textbf{1.0}    &   \textbf{1.0}   &   \textbf{1.0}    &   \textbf{1.0}    &  \textbf{1.0}    &  \textbf{1.0} \\
\hline
DPC-CE   &    \textbf{1.0}    &   \textbf{1.0}   &   \textbf{1.0}    &   \textbf{1.0}    &  \textbf{1.0}    &  \textbf{1.0} \\
\hline
Our Method  &   \textbf{1.0}    &   \textbf{1.0}   &   \textbf{1.0}    &   \textbf{1.0}    &  \textbf{1.0}    &  \textbf{1.0} \\
\hline

\multicolumn{1}{|c|}{\multirow{2}*{}}&\multicolumn{3}{c|}{Threecircles}&\multicolumn{3}{c|}{Jain}\\
\cline{2-7}
~ & FMI & ARI & NMI & FMI & ARI & NMI
\\
\hline
K-means    &    0.4045 & 0.0555  &  0.1637 & 0.7005  & 0.3241 & 0.3690\\
\hline
DPC    &    0.5161 & 0.2514    &   0.3703 & 0.8779  & 0.7055 & 0.6447\\
\hline
DBSCAN    &    0.9193    &   0.8739   &   0.8647    &   0.9767   &  0.9473    &  0.8930 \\
\hline
SNN-DPC    &    0.7160    &   0.5310   &   0.6860    &   \textbf{1.0}    &  \textbf{1.0}    &  \textbf{1.0} \\
\hline
DGDPC    &    \textbf{1.0}    &   \textbf{1.0}   &   \textbf{1.0}    &   \textbf{1.0}    &  \textbf{1.0}    &  \textbf{1.0} \\
\hline
DPC-CE   &    \textbf{1.0}    &   \textbf{1.0}   &   \textbf{1.0}    &   \textbf{1.0}    &  \textbf{1.0}    &  \textbf{1.0} \\
\hline
Our Method  &   \textbf{1.0}    &   \textbf{1.0}   &   \textbf{1.0}    &   \textbf{1.0}    &  \textbf{1.0}    &  \textbf{1.0} \\
\hline

\multicolumn{1}{|c|}{\multirow{2}*{}}&\multicolumn{3}{c|}{Fourlines}&\multicolumn{3}{c|}{Unbalance}\\
\cline{2-7}
~ & FMI & ARI & NMI & FMI & ARI & NMI
\\
\hline
K-means    &    0.6462 & 0.5024    &    0.6725 & 0.8142  & 0.8463 & 0.8107\\
\hline
DPC    &    0.7850 & 0.7115    &   0.7698 & \textbf{1.0}  & \textbf{1.0} & \textbf{1.0}\\
\hline
DBSCAN    &    \textbf{1.0}    &   \textbf{1.0}   &   \textbf{1.0}    &   \textbf{1.0}    &  \textbf{1.0}    &  \textbf{1.0} \\
\hline
SNN-DPC    &    \textbf{1.0}    &   \textbf{1.0}   &   \textbf{1.0}    &   \textbf{1.0}    &  \textbf{1.0}    &  \textbf{1.0} \\
\hline
DGDPC    &    \textbf{1.0}    &   \textbf{1.0}   &   \textbf{1.0}    &   \textbf{1.0}    &  \textbf{1.0}    &  \textbf{1.0} \\
\hline
DPC-CE   &    \textbf{1.0}    &   \textbf{1.0}   &   \textbf{1.0}    &   \textbf{1.0}    &  \textbf{1.0}    &  \textbf{1.0} \\
\hline
Our Method  &   \textbf{1.0}    &   \textbf{1.0}   &   \textbf{1.0}    &   0.9999    &  0.9999    & 0.9994 \\
\hline

\end{tabular}
\end{table}

\subsection{4.2 Experimental results on real world data sets}

\begin{figure}
\includegraphics[width=\columnwidth]{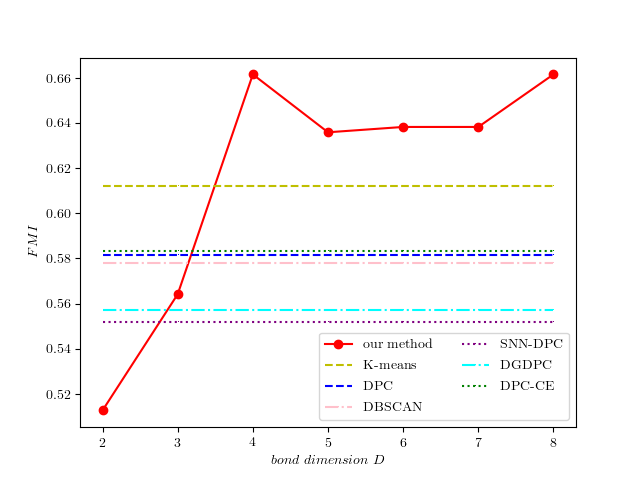} 
\caption{The FMI of the Wine data set using
our method and other baseline models. Bond dimension D determines the number of parameters in the MPS.}
\label{fig:fig7} 
\end{figure}

\begin{figure}
\includegraphics[width=\columnwidth]{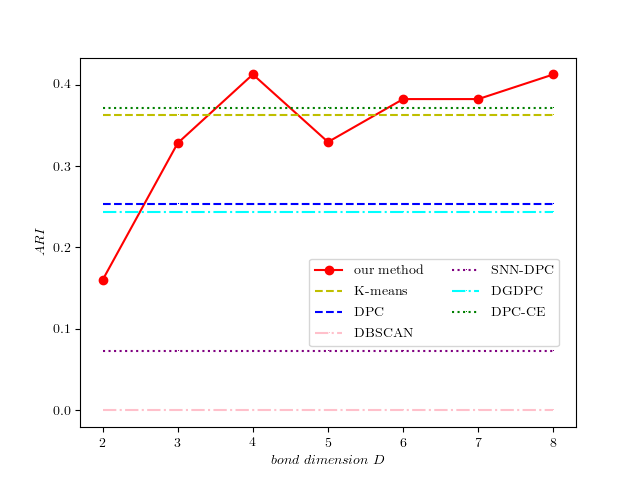} 
\caption{The ARI of the Wine data set using
our method and other baseline models. Bond dimension D determines the number of parameters in the MPS..}
\label{fig:fig8} 
\end{figure}

\begin{figure}
\includegraphics[width=\columnwidth]{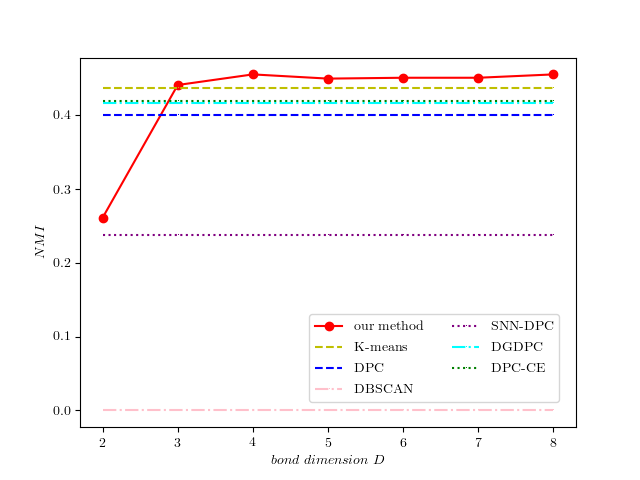} 
\caption{The NMI of the Wine data set using
our method and other baseline models. Bond dimension D determines the number of parameters in the MPS.}
\label{fig:fig9} 
\end{figure}

In this section, experiments are validated on 4 real world data sets. Details for these data sets are given in Table ~\ref{tab:tab4}, and parameter information is also listed in Table ~\ref{tab:tab2}. According to the entanglement entropy area law, in the quantum state represented by MPS, the bond dimension D determines the upper limit of the entanglement entropy it can accommodate, and the parameter space of the MPS grows at the scale of $D^2$. Therefore, taking the Wine data set as an example, we compared the clustering performance under different bond dimension D, which is shown in FIG.~\ref{fig:fig7}-~\ref{fig:fig9}. As can be seen from the figure, when $D\geq 4$, our algorithm has a significant improvement in the value of FMI compared to other algorithms. And at $D\geq 3$, our algorithm already exceeds the performance of other algorithms on NMI. We believe that the quality of the clustering results has a great relationship with the entanglement entropy of the trained MPS. The larger the entanglement entropy means the smaller the local entanglement of the data, the stronger the expressability of the MPS to the data. More details can be found in the Appendix.

We still use the three metrics of NMI, ARI, and FMI to compare the clustering results with other methods. In Table ~\ref{tab:tab5} we present the results of our algorithm when $D=8$. It can be seen that the performance of our three indicators is currently the best on the Wine data set, and our FMI outperforms other methods on the Vehicle and Yeast data sets. On the Abalone data set, its FMI and NMI are the highest among these methods, and our algorithm are only 0.0210, 0.0103 and 0.0660 lower than SNN-DPC in FMI, ARI, and NMI, respectively. In terms of ARI, the result obtained by our algorithm is 0.0416, close to the largest one (DPC-CE’s ARI = 0.0613).


All in all, the clustering results are encouraging, and they show that the tensor network clustering algorithm can achieve better results than other existing algorithms even when the number of clusters is not known in advance. It demonstrates the excellent ability of our algorithm to handle real-world data sets.

\begin{table}
\caption{Description of the real world data set.} 
\label{tab:tab4}
\begin{tabular}{| c | c | c | c | }
\hline                              
Data set & Point & Attribute & Clusters \\
\hline
Wine & 178 & 13 & 3 \\
Vehicle & 846 & 18 & 4\\
Yeast & 1484 & 9 & 10 \\
Abalone & 4177 & 8 & 29 \\
\hline
\end{tabular}
\end{table}

\begin{table}[!htbp]
\caption{Clustering results of all methods in real world data set, like before, the bond dimension in MPS is set to be 8.}
\label{tab:tab5}
\centering
\begin{tabular}{ | c | c | c | c | c | c | c |  }
\hline

\multicolumn{1}{|c|}{\multirow{2}*{Method}}&\multicolumn{3}{c|}{Wine}&\multicolumn{3}{c|}{Vehicle}\\
\cline{2-7}
~ & FMI & ARI & NMI & FMI & ARI & NMI
\\
\hline
K-means    &    0.5835 & 0.3711  &  0.4288  & 0.3590  & 0.1216 & 0.1867\\
\hline
DPC    &    0.5817 & 0.2535 & 0.3997 & 0.3973 & 0.0829 & 0.1136\\
\hline
DBSCAN    &   0.5782 & 0 & 0 & 0.4873 & 0 & 0 \\
\hline
SNN-DPC    & 0.5520 & 0.0728 & 0.2375 & 0.3620 & 0.0532 & 0.0589 \\
\hline
DGDPC    & 0.5570 & 0.2436 & 0.4169 & 0.4361 & \textbf{0.4982} & \textbf{0.4733}\\
\hline
DPC-CE    &  0.5834 & 0.3715 & 0.4193 & 0.4254 & 0.4020 & 0.4020\\
\hline
Our Method &  \textbf{0.6616} & \textbf{0.4125} & \textbf{0.4550} & \textbf{0.4951} & 0.0006 & 0.0156\\
\hline

\multicolumn{1}{|c|}{\multirow{2}*{}}&\multicolumn{3}{c|}{Yeast}&\multicolumn{3}{c|}{Abalone}\\
\cline{2-7}
~ & FMI & ARI & NMI & FMI & ARI & NMI
\\
\hline
K-means    &    0.2980 & \textbf{0.1331}  &  \textbf{0.2436}  & 0.1119  & 0.0433 & 0.1611\\
\hline
DPC    &    0.4703 & 0.0107 & 0.1224 & 0.1356 & 0.0348 & 0.0432\\
\hline
DBSCAN    &   0.4037 & 0.0254 & 0.0296  & 0.2248 & 0.0385 & 0.0980\\
\hline
SNN-DPC    & 0.4422 & 0.0121 & 0.1182 & \textbf{0.2459} & 0.0519 & \textbf{0.1685}\\
\hline
DGDPC    & 0.4397 & 0.0987 & 0.1226 & 0.1937 & 0.0529 & 0.1066\\
\hline
DPC-CE    &  0.4705 & 0.1185 & 0.1277 & 0.2250 & \textbf{0.0613} & 0.1373\\
\hline
Our Method &  \textbf{0.4710} & 0.0125 & 0.0653 & 0.2249 & 0.0416 & 0.1025\\
\hline

\end{tabular}
\end{table}

\subsection{4.3 Experimental results on image data sets}
Finally, we compare our algorithm with other good clustering algorithms on computer vision benchmark data sets to demonstrate the effectiveness of our algorithm. Their statistical information is shown in Table ~\ref{tab:tab6}. Both the MNIST and Fashion data sets have 70,000 images, each containing 28*28 grayscale pixels. The USPS data set has a relatively small number of images, with 9298 images, each containing 16*16 grayscale pixels. Like the method we deal with in the above, we use max-min normalization to preprocess the data, which performs a linear transformation on the original data. 

When data with high-dimensional feature space is involved in practical applications, a series of preprocessing steps are needed in order to obtain better clustering, considering the time cost. Here we first use the autoencoder to reduce the dimension of the data. The autoencoder consists of two parts. The first part is the encoder $E$, which compresses the initial data $x$ to the latent space through a learned feature vector $e = f(x)$, and the second part is called decoding $D$, which learns a new function $g$ that maps the compressed data into the original feature space. The training process of the autoencoder can be expressed as:

\begin{align}
\mathop{argmin} \limits_{(e,d)\in E\times D} \epsilon(x, g(f(x)))
\end{align}

where $\epsilon(x, g(f(x)))$ is the reconstruction error between the input data $x$ and $g(f(x))$.

Although autoencoders are a common and efficient way to compress data. But it does not preserve the distance information between data well enough. With this in mind, we need to take a further approach to the data obtained from the autoencoder.

\begin{figure}
\includegraphics[width=\columnwidth]{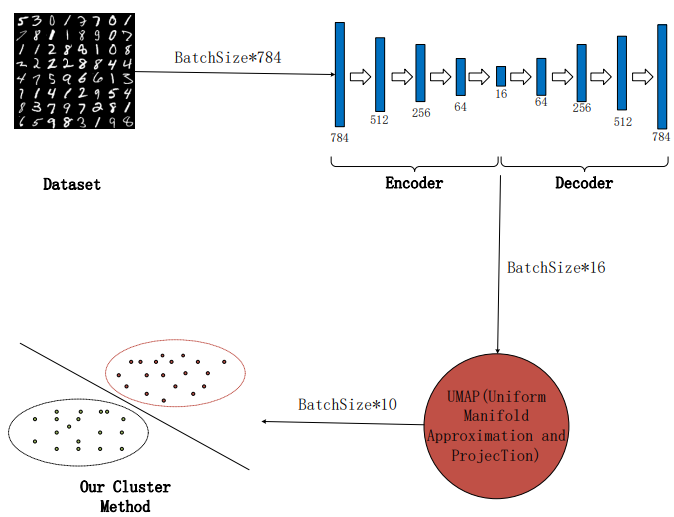} 
\caption{A flowchart of the structure when processing computer vision data sets. The dimensions of the data after each step of preprocessing are reflected on the connection line. The number of neurons in each layer of the autoencoder is shown below it.}
\label{fig:fig12} 
\end{figure}

UMAP (Uniform Manifold Approximation and Projection)~\cite{mcinnes2018umap} is a common dimensionality reduction technique that can be used for general nonlinear dimensionality reduction, but also for t-SNE-like visualizations. UMAP is a dimensionality reduction algorithm based on manifold learning technology and topological data analysis ideas. It is divided into two steps, the first step is to learn the manifold structure of the data in the high-dimensional space, and the second step is to find the low-dimensional representation of this manifold. Compared with t-SNE, it significantly improves the speed and better preserves the global structure of the data. Therefore, in practice, we will first use the autoencoder to act on the original data to learn an initial representation, then we relearn the data from the autoencoder by searching for a more clustered manifold using a local distance-preserving manifold learning method. Here, the structure of the autoencoder we use is $FC_{512}\rightarrow FC_{256}\rightarrow FC_{64}\rightarrow FC_{16}\rightarrow FC_{64}\rightarrow FC_{256}\rightarrow FC_{512}$. Where $FC_{512}$ indicates that it is a fully connected layer with 512 neurons. This means that with this autoencoder, we compress the original data into a 16-dimensional latent space. Next, the 16-dimensional data is further reduced to 10-dimensional using the UMAP method, which is then fed into our tensor network clustering algorithm. The details of the structure can be seen in Fig.\ref{fig:fig12}.
 
Following the above works, we use two metrics, ACC and NMI, to evaluate the performance of the algorithm. In Table ~\ref{tab:tab7}, we present a comparison of our method with other top-performing algorithms. Among all the algorithms, only DBSCAN, DPC, DED, DDC and our method do not need to know the number of clusters in advance, While other algorithms take real clusters numbers as known conditional inputs. 

It can be seen that in the algorithm with unknown number of clusters, our algorithm demonstrates the current state-of the-art result on the MNIST and USPS data sets. Although the ACC is a little lower than the DDC algorithm on the Fashion data set, but our results are better than the DDC algorithm on the NMI indicator. Even compared with the state-of-the-art clustering algorithms~\cite{PhysRevA.105.052424}, our algorithm is only 1.24\% and 3.22\% lower in ACC and NMI on the MNIST data set, which provides a competitive scheme.

\begin{table}
\caption{Description of the common used image data set.} 
\label{tab:tab6}
\begin{tabular}{| c | c | c | c | }
\hline                              
Data set & Point & Attribute & Clusters \\
\hline
MNIST & 70000 & 784 & 10 \\
Fashion & 70000 & 784 & 10 \\
USPS & 9298 & 256 & 10 \\
\hline
\end{tabular}
\end{table}

\begin{table}[!htbp]
\caption{ACC and NMI of our method compared to other top-performing image clustering
models. The best results are stressed in bold and the second best results are stressed in Italian. }
\label{tab:tab7}
\centering
\begin{tabular}{ | c | c | c | c | c | c | c |}
\hline
\multicolumn{1}{|c|}{\multirow{2}*{Method}}&\multicolumn{2}{c|}{MNIST}&\multicolumn{2}{c|}{USPS}&\multicolumn{2}{c|}{Fashion}\\
\cline{2-7}
~ & ACC & NMI & ACC & NMI & ACC & NMI \\
\hline
K-means    &   53.91    &   49.04    &   65.76    &   60.98    &   52.22    &   51.1  \\
DBSCAN & - & - & 16.7 & 0 & 10.0 & 0 \\
DPC & - & - &  39.0 & 43.3  &  34.4 & 39.8  \\
DEC~\cite{pmlr-v48-xieb16}    &   84.3    &   83.4    &   76.2    &   76.7    &   51.8    &   54.6\\
IDEC~\cite{guo2017improved} & 88.06 & 86.72  &  76.05  &  78.46 &  52.9 &   55.7 \\
JULE~\cite{yang2016joint}    &   96.4    &   91.3    &  95.0    &   91.3    &   -    &   -  \\
DEPICT~\cite{Dizaji_2017_ICCV}    &   96.5    &   91.7    &   96.4    &   \textit{92.7}    &   39.2    &   39.2\\
EnSC~\cite{you2016oracle}    &   96.3    &   91.5    &   61.0    &   68.4    &   62.9    &   63.6 \\
InfoGAN~\cite{chen2016infogan}   &   87.0    &   84.0    &   -    &  -    &   61.0    &   59.0 \\
ClusterGAN~\cite{Mukherjee_Asnani_Lin_Kannan_2019}   &   95.0    &   89.0    &   -    &  -    &   63.0    &   64.0\\
DualAE~\cite{Yang_2019_CVPR}   &   \textit{97.8}    &   \textit{94.1}    &   86.9    &  85.7    &   \textbf{66.2}    &   64.5 \\
ConvDEC~\cite{guo2018deep}    &   94.0    &   91.6    &   78.4    &   82.0    &   51.4   &  58.8 \\
DDC~\cite{ren2020deep}    &   96.5    &   93.2    &   96.7  &   91.8    &   61.9   &   68.2 \\
ADSSC-MPS-8~\cite{PhysRevA.105.052424} & \textit{99.04}    &   \textit{97.22}    &   \textit{98.82}    &   \textit{96.62}    &   \textit{65.61}    &   \textbf{72.15} \\
\hline
MPS-8    &   \textbf{97.8}   &   \textbf{94.0}    &   \textbf{97.1}    &   \textbf{92.6}    &   59.99   &   \textit{68.49} \\

\hline
\end{tabular}
\end{table}

\section{5. Discussion}
In this paper, we introduce a density-based clustering algorithm with tensor networks. It is an efficient quantum-inspired unsupervised learning algorithm and can recognize clusters of arbitrary shape and size. We have introduced the main algorithm, and in order to demonstrate its effectiveness, we apply it on various types of data sets, including synthetic data sets, real world data sets, and computer vision data sets. It should be emphasized that when calculating the density, it converts the input data into a matrix product state with a bond dimension equal to 1, connects the data of each dimension in the data set through quantum entanglement, and compares the MPS under different bond dimension D. On this exponentially large Hilbert space, cluster centers can be found better, and our algorithm can achieve good results even when D is small. It can also be seen that larger quantum entanglement tends to provide better clustering results.

Although our method proves to be a simple and effective clustering algorithm for tensor networks, it also suffers from some problems. The first is that its adaptability is relatively poor. Although the value of $d_c$ is the same in many data sets, the value of $d_f$ is not the same. That is, our parameters are not the same for different data sets. So how to adaptively select parameters will be one of the future work. In addition, it is worth mentioning that MPS is the simplest tensor network structure, in addition to the Tree Tensor Network (TTN), projected entangled pair states (PEPS), the String Bond States (SBS), etc.. They are more difficult to maneuver than MPS, but more effective. So it would be an interesting idea to use other tensor network results to do natural clustering of the data. Besides, when dealing with computer vision data sets, we use classical dimensionality reduction methods, so it is also a promising attempt to combine tensor networks and machine learning for data dimensionality reduction.



\section{ACKNOWLEDGMENTS}
Y. Shang thanks the support of National Natural Science Foundation of China (Grant No. 61872352) and the Program
for Creative Research Group of the National Natural Science Foundation of China (Grant No. 61621003).

\section{APPENDIX: Entanglement entropy of the MPS}
Entanglement entropy is a fundamental concept in quantum physics that characterizes the entanglement present in a quantum system. It is obtained by dividing the system into two subsystems and measuring the amount of entanglement between them. The larger the entanglement entropy, the smaller the local correlations within the data and vice versa. One important property of MPS is that they obey an area law for entanglement entropy, which means that the entanglement entropy of an MPS with bond dimension $\chi$ and length $m$ is upper bounded by a value that is independent of the system length. This property has significant implications for the computational complexity of simulating and manipulating MPS, as well as for the behavior of quantum many-body systems in general.

For a center-normalized MPS, the entanglement entropy reaches its maximum value when the Schmidt coefficients are all equal to $1/\sqrt\chi$ \cite{RevModPhys.82.277}. This can be seen by dividing the MPS into two subsystems at its orthogonal center and using Schmidt decomposition to determine the entanglement entropy.

Take the Wine dataset as an example. We consider the canonical form of the MPS trained on all data in the dataset to obtain the normalized entanglement spectrum. We convert it to the center canonical form at the ${\lfloor m/2 \rfloor}$ bond and split it into left and right parts using Schmidt decomposition

\begin{align}
\Phi^{\vec\tau}=\sum_{i=1}^r\lambda_i |\phi_i^L\rangle\otimes|\phi_i^R\rangle
\end{align}

\begin{figure}
\includegraphics[width=\columnwidth]{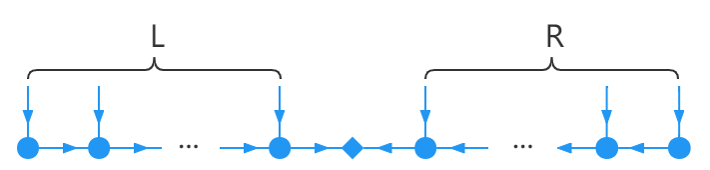} 
\caption{The specific structural form of MPS. The MPS is divided into two subsystems L and R from the middle to calculate the entanglement entropy between them. Among them, the subsystem L satisfies the left canonical form, and the subsystem R satisfies the right canonical form.}
\label{fig:fig10} 
\end{figure}

Where $r$ is the Schmidt rank, $\lambda_i$ is the Schmidt coefficient, which is a non-negative real number and satisfies $\sum_{i=1}^r\lambda_i^2=1$. As shown in Fig.\ref{fig:fig10}. Therefore, its entanglement entropy S can be obtained in the following

\begin{align}
    S=-\sum_{i=1}^r \lambda_i^2 ln (\lambda_i^2)
\end{align}

\begin{figure}
\includegraphics[width=\columnwidth]{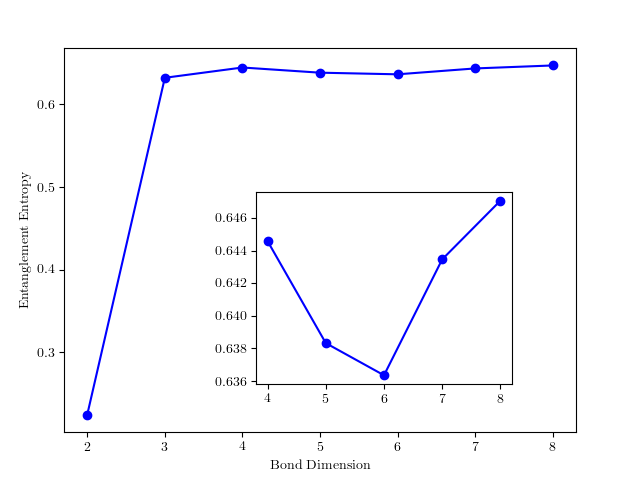} 
\caption{Taking the Wine data set as an example, the entanglement entropy measured at the ${\lfloor m/2 \rfloor}$ bond. It compares the entanglement entropy MPS can accommodate under different bond dimensions.}
\label{fig:fig11} 
\end{figure}

The result is summarized in Fig.~\ref{fig:fig11}. It shows that the entanglement entropy is relatively large when bond dimension equals to 4 and 8, which is consistent with the clustering results that can work better under the corresponding bond. And its trend is basically the same as that of FMI, ARI and NMI. It can be seen that the final clustering result has an important relationship with the entanglement entropy of the trained MPS. Also, compared with other classical algorithms, it can be seen that establishing entanglement through quantum methods is more conducive to finding hidden relationships between data. The strength of nonlocality has been shown to impact the performance of the clustering algorithm in this example, with stronger nonlocality leading to improved results.

\bibliographystyle{apsrev4-1}
%

\end{document}